# USING OBJECTIVE WORDS IN THE REVIEWS TO IMPROVE THE COLLOQUIAL ARABIC SENTIMENT ANALYSIS

Omar Al-Harbi

Department of Computer and Information, Community College, Jazan University, Jazan, KSA

## ABSTRACT

*One of the main difficulties in sentiment analysis of the Arabic language is the presence of the colloquialism. In this paper, we examine the effect of using objective words in conjunction with sentimental words on sentiment classification for the colloquial Arabic reviews, specifically Jordanian colloquial reviews. The reviews often include both sentimental and objective words; however, the most existing sentiment analysis models ignore the objective words as they are considered useless. In this work, we created tow lexicons: the first includes the colloquial sentimental words and compound phrases, while the other contains the objective words associated with values of sentiment tendency based on a particular estimation method. We used these lexicons to extract sentiment features that would be training input to the Support Vector Machines (SVM) to classify the sentiment polarity of the reviews. The reviews dataset have been collected manually from JEERAN website. The results of the experiments show that the proposed approach improves the polarity classification in comparison to two baseline models, with accuracy 95.6%.*

## KEYWORDS

*Arabic sentiment analysis, opinion mining, colloquial Arabic language, colloquial Jordanian reviews*

## 1. INTRODUCTION

Measuring the satisfaction and obtaining the feedback from users have always been the concern of companies that offer services or products to make decisions that would improve their business. Years ago, this process was hard, but with the advancement in web 2.0 platforms such as forums, blogs, and social media including their extensive data of opinions made that easier [1]. Platforms that allow the customers to express their opinions and emotions in reviews about the products have become a reliable source of feedback to both companies and customers. These reviews can make the process of satisfaction detection for companies easier and can influence on customer's decision making whether to buy a product [2]. However, the vast amount of reviews that are published in digital form about products makes it harder for companies and customers to find the true underlying sentiment about the products without an automatic mechanism. Therefore, there was a need for systems that can automatically perform the process of sentiment analysis or opinion mining in the reviews.

Sentiment analysis or opinion mining field is a task of Natural Language Processing (NLP) which aims to automatically detect the subjective information in textual information and predict its sentiment polarity [3]. Recently, the attention is being paid to the sentiment analysis and its applications in different domains such as finance, economic, healthcare, etc. The core function of sentiment analysis process is assigning positive, negative, or neutral polarities to opinionated texts [4]. In general, sentiment analysis has been investigated at three levels [5]: document level, sentence level, and entity and Aspect level. The approaches of sentiment classification that have been used in literature can be classified into two primary categories: supervised approach and semantic orientation approach [6]. The former is also known as a corpus-based approach which uses machine learning algorithms to classify the sentiment into binary or multiple classes. In this approach, there must be annotated data from which a set of features is extracted as a training data used by a classifier to build a model for predicting the classes of a testing data using one of the machine learning algorithms like SVM, Naive Bayesian, K-Nearest Neighbor, etc. The later is also known as lexicon-based approach in which sentiment lexicons and other linguistics resources are used to classify the sentiment polarity. In this approach, the sentiment information is extracted and represented by numeric values to be summed up in a value indicating the sentiment polarity of a sentence.

Each approach has its advantages and disadvantages, however, studies such as [7, 8] show that the two approaches have complementary performances. For example, Yang et al. [8] illustrate how the combination of the two approaches can resolve the problem of the learning classifiers which is being conservative in classifying instances as positive because positive reviews usually contain many objective statements. In contrast, the lexicon-approaches tend to classify negative or neutral instances as positive when there are a few positive words appear in the document. In this research, we incorporated lexicons into a machine learning classifier. The lexicons used to extract different sentiment features that can improve the performance of the learning classifier for colloquial Arabic language.

The research in sentiment analysis of English language has achieved considerable progress, whereas it is still limited in the Arabic language. One of the most challenging issues in the Arabic language is the presence of colloquialism, as there are no specific rules that govern the colloquial Arabic. In sentiment analysis literature, the MSA is widely researched and several tools and resources were developed, whereas there are fewer researches concern with colloquial Arabic [9, 10, 11, 12] . Additionally, different approaches were adopted and adapted whether for MSA or colloquial Arabic. However, to the best of our knowledge, none investigated the effect of objective words on the sentiment classification. The researchers usually adopt the suggestion of Pang and Lee [13] that refers to excluding the objective texts would improve the sentiment classification.

In this work, we investigate the effect of objective words on sentiment classification for colloquial Arabic reviews, specifically Jordanian colloquial reviews. To the best of the author knowledge, there is no much research concern with colloquial Jordanian in sentiment analysis other than [14, 11, 15] . Thus, we introduce a new approach that incorporates different lexicons into SVM classifier to classify the reviews into either positive or negative class. The lexicons include colloquial sentiment words, objective words that have sentiment tendency, and sentiment compound phrases to extract a set of sentiment features that can improve the classifier. Furthermore, we investigated the effect of combining the sentiment features with different n-gram models on the classifier performance.

The paper is conducted as follows. Section 2 introduces the background of the Arabic language and colloquial dialects. Section 3 presents related work. Section 4 introduces the proposed approach. Section 5 discusses experimentations and results. Finally, Section 6 concludes this work.

## 2. BACKGROUND OF THE ARABIC LANGUAGE AND COLLOQUIAL DIALECTS

The Arabic language is one of the Semitic languages, and used by about 325 million native speakers to
daily communication [16]. It is also one of the languages in United Nations as are English, and French.

The Arabic script is the second most familiar script in the world after Latin [17]. It is used in Arabic and other languages such as Ottoman Turkish, Persian, Urdu, Afghan, and Malay. The Arabic language has a morphologically complex style that has a high inflectional and derivational nature [18]. This nature would affect the linguistic features such as, gender, number, tense, person and etc [19]. Additionally, this nature would makes Arabic language the richest with vocabularies compared to other natural languages [20].

The Arabic language is a collection of variants among which the MSA variant is the most used [21]. MSA is the most common and understood by all over Arabic world, and used in books, newspapers, news, formal speeches, subtitles, etc. The MSA derived from the classical Arabic language, and they have several features in common. However, they are treated separately and have differences in aspects such as lexicon, stylistics, and certain innovations on the periphery [22]. The classical Arabic language has remained unchanged, intelligible and functional for more than fifteen centuries [23]. It is the written language of the Quran, and it is used by around 1.4 billion Muslim to perform their daily prayers. Recently, the Arabic language has attracted the attention of natural language processing researchers. Different state-of-the-art systems have been developed for different application including sentiment analysis. However, these applications had to deal with several challenging issues relevant to the Arabic language; one of them is the colloquialism.

Colloquial Arabic dialects are also rooted in classical Arabic and MSA, and the script is the same. However, there is a great variety in Arabic dialects among the Arab countries or even different regions at the same country. For example, Table 1 illustrates how the sentence (ماذا تريد؟) which means (What do you want?) is differently written with different colloquial dialects.

Table 1: Variations among different colloquial Arabic dialects.

| Colloquial dialect | Sentence | Buckwalter |
|---|---|---|
| Jordanian | أيش بدك | >y$ bdk |
| Egyptian | عايز ايه | EAyz Ayh |
| Saudi | وش تبغى | w$ tbgY |
| Tunisian | شنو تحب | $nw tHb |
| Algerian | واش تحب | wA$ tHb |

Obviously, there are no standard rules for colloquial dialects at the levels of orthography, morphology, phonology, and lexicon. The variation of colloquial Arabic dialects can be on different dimensions, Habash [21] mentioned two major dimensions: geography and social classes. Based on geography, the colloquial dialects categorized as Egyptian, Levantine, Gulf, North African, and Iraqi. Socially, the dialects can be classified into three categories: urban, rural, and bedouin. Shaalan et al [9] pointed out that the differences between dialects and MSA because behaviors such as replacing characters and change the pronunciation or the style of writing of nouns, verbs, and pronouns. Consequently, new colloquial words will continue appearing, and the gap between MSA and dialects will increase. In this case, according to [24, 23], applying the tools of MSA to the most colloquial texts will give results far from accurate since there are variances in grammar syntax, and expressions. such as lexicons, annotated corpora, and parsers. For example, people use new and different ways to express their sentiment such as transliterated English like (كيوت ،نايس ،لول) which means (LoL, nice, cute), and newly created compound phrases like (سعرو فيه) which means (worthy). Because such challenges, processing Arabic dialects in sentiment analysis is difficult, and most researchers prefer to deal with MSA texts, since MSA was robustly researched and have a considerable amount of resources.

The colloquial Jordanian dialect is spoken by more than 6 million. According to (Cleveland 1963), the colloquial Jordanian has three categories. First, the urban dialect which has emerged as a result of internal and external migrations to the main cities. Second, the rural dialect which is often spoken in villages and small cities, and it has two categories; Horan dialect which is used in the area north and west Amman, and Moab dialect which is used in the area of South Amman. Third, Bedouin dialect which is spoken by Jordanian Bedouins who live in the desert, and is not common in the urban and rural regions. Table 2 shows examples of how the colloquial Jordanian dialect varies in a sentence like (ما خطبه؟) which means (What is the matter with him?).The migration has played a significant role in the formation of Jordanian dialect. Since 1984, Jordan has received a considerable amount of Palestinian refugees who settled all over the region. The contact between Palestinians and Jordanians has created new and complex patterns of dialects [25]. Furthermore, a flood of Syrian refugees recently was accepted, that made Jordan dialect observably propagated. Based on the introduced facts, Jordanian dialect continuously adds new suffixes, prefixes, and clitics that would generate new words, stop-words, contrary words, and negation words (e.g. مش، مهوش، منو، خَروح، خرمان).

Table 2: Categories of colloquial Jordanian dialects.

| Colloquial Jordanian | Sentence | Buckwalter |
|---|---|---|
| Urban | ماله هاد | mAlh hAd |
| Rural | مالو هاظ | mAlw hAZ |
| Bedouin | علامو هاذ | ElAmw hA* |

## 3. RELATEDWORK

Different studies investigated the effect of using objective words on sentiment classification for English language. Hung and Lin [26] pointed to the fact that more than 90% of the words in SentiWordNet are objective words. These objective words are ignored in the most existing sentiment models, because they are considered useless. However, in their work, they used the objective words in the sentiment classification after reevaluating their sentiment tendency based on the presence of the words in positive and negative reviews. The results of the experiment have shown an improving effect on the performance accuracy with 4.10% compared to using non-revised SentWordNet. Another work of Kaewpitakkun et al. [27], also found that reevaluating the objective words in SentiWordNet would improve the accuracy of sentiment classification. The same conclusion was reached by Amiri and Chua [28] and Ghang and Shah [29] in their work.

Up to now, most researchers on Arabic sentiment analysis have targeted MSA form, due to the availability of resources and tools. Whereas, we found few related work on colloquial Arabic sentiment analysis. The most of the related work considered the colloquial Egyptian such as [10, 30, 31, 32, 33], since it is the most widely spoken colloquial dialect in the Middle East by more than 80 million people. In this section, we present some previous work that considered MSA and colloquial Arabic in sentiment analysis.

Shoukry and Rafea [31] proposed a hybrid approach for sentiment classification of colloquial Egyptian tweets. They used the SVM with different types of features such as n-grams, and sentiment scores that obtained based on sentiment words, and emoticons. In their work, the preprocessing includes character normalization, stemming, and stop word removal. They manually built the lexicons that used in the process of classification. To train the classifier, they used 4800 tweets of which 1600 are positive, 1600 negative, and 1600 neutral. The objective of their work was examining the effect of the corpus sizes on the Machine learning classifier, and the effect of their proposed hybrid approach. Based on the reported results, there was improving effect on sentiment classification.

Ibrahim et al. [30] used a semi-supervised approach for sentiment analysis of MSA and colloquial Egyptian. They introduced a high coverage Arabic sentiment lexicon with 5244 terms, and a lexicon of idioms/saying phrases with 12785 phrases. Regarding feature selection, they extracted different linguistic features to improve the classification process. For classification, they used the SVM technique. Their dataset consists of 2000 statement divided into 1000 tweet and 1000 microblogging reviews. The reported accuracy of the SVM classifier was 95%.

Azmi and Alzanin [34] introduced Aara'which is a mining system for public comments written in colloquial Saudi. They employed the Naive Bayes algorithm with a revised n-gram approach for classi_cation. The dataset consists of 815 comments which were gathered manually from online newspapers, and then split into a training set and testing set. The accuracy of the system was 82%.

Salamah and Elkhli_ [35] proposed an approach for sentiment classification of colloquial Kuwaiti in microblogging. The approach employed a lexicon of colloquial Kuwaiti adjectives, nouns, verbs, and adverbs. They tested their approach on a manually annotated corpus comprised of 340,000 tweets. For classification, they used SVM, J48, ADTREE, and Random Tree classifiers. The approach yielded the best results using SVM with a precision and recall of 76% and 61% respectively.

Abdul-Maged et al. [10] presented SAMAR for subjectivity and sentiment analysis for Arabic social media reviews. In this work, they considered both MSA and colloquial Egyptian. In this work, different features were used include author information, stemming, POS tagging, dialect and morphology features. For classification, they used SVM classifier over a variety datasets. Concerning colloquial Arabic, they noticed that the presence of colloquial tweets would affect the SSA negatively since the most tweets are subjective and negative in sentiment. The highest accuracy reported through the colloquial-specific sentiment experiments was 73.49%.

The work of Mourad and Darwish [36] focused on Subjectivity and Sentiment Analysis (SAA) on Arabic news articles and dialectal Arabic microblogs from Twitter. A random graph walk approach was employed to expand the Arabic SSA lexicon using Arabic-English phrase tables. They used two classifiers in the experiments, the NB and SVM classifiers with features such as stem-level features, sentence-level features, and positive-negative emoticons. The accuracy was 80% for news domain and 72.5% for tweets.

Duwairi [15] introduced a framework for sentiment analysis of Arabic tweets with the presence of colloquial Jordanian. The approach utilizes machine learning classifier and colloquial lexicon which maps colloquial words to their corresponding MSA words. 22550 tweets were collected using Twitter API and annotated using a crowd sourcing tool. In this work, utilizing the colloquial lexicon achieved a slight improvement. Two classifiers were used to determine the polarity, namely: NB and SVM, the F-measure of the two classifiers was 87.6% and 86.7% respectively.

Finally, Abdulla et al. [14] presented a lexicon-based approach for analyzing opinions written in both MSA and colloquial Jordanian. The lexicon size was 3479 words, and the dataset composed of 2000 tweet were collected and manually annotated. For feature extraction, they used unigram technique, and then they used an aggregation tool to calculate the weights of tweets to generate the polarity. They performed a comparison between lexicon-based and corpus-based approaches; as noticed from the results corpus-based approach remarkably outperformed the lexicon-based approach. The final reported accuracy of lexicon based approach was 59.6%.

As we note, different approaches, methods, resources, and colloquial dialects have been researched in the context of Arabic sentiment analysis. However, to the best of the author knowledge, none examined the effect of using the objective words in the sentiment classification either in MSA or colloquial Arabic. The researchers usually adopt the suggestion of Pang and Lee [13] that refers to excluding the objective texts would improve the sentiment classification. In this work, we decided to examine the effect of the objective words by assigning sentiment tendency for them based on a method will be discussed in section 3. Unfortunately, we found few work that concern with colloquial Jordanian, and we have not found public datasets to be used and compared with the proposed approach. Thus, we manually collected and annotated our own datasets reviews, and then we built new sentiment lexicons of objective and opinionated colloquial words and phrases.

# 4. THE PROPOSED APPROACH

This section presents the proposed approach for classifying the sentiment polarity of colloquial Jordanian reviews. In this research, we used an approach that employs different lexicons for extracting sentiment features from the reviews to be fed to the SVM classifier. The core of this approach is to combine a colloquial sentiment words lexicon with a lexicon of objective words that have sentiment tendency, to be involved in the process of sentiment polarity classification. The framework of the proposed approach is illustrated in Fig 1.

## 4.1. DATA COLLECTION AND ANNOTATION

An annotated dataset is required to train the classifier. To the best of our knowledge, there is no publicly available corpus for Jordanian colloquial dialect. Thus, we manually built our corpus consisting 2730 reviews of which 1527 were positive, and 1203 were negative. The data is collected from JEERAN1 website which is a platform for user's reviews about places, services, and products in Jordan. The corpus consists of MSA and colloquial Jordanian reviews about various domains (restaurants, shopping, fashion, education, entertainment, hotels, motors, and tourism). This corpus is mostly written by reviewers from the public and consists of short and long reviews. Two Jordanian native speakers annotated the polarity of the reviews, and a good agreement was reflected. In this work, only the positive and negative reviews were considered, while the reviews such as neutral, sarcastic, and uncertain have been disregarded in this work.

1http://jo.jeeran.com/amman/

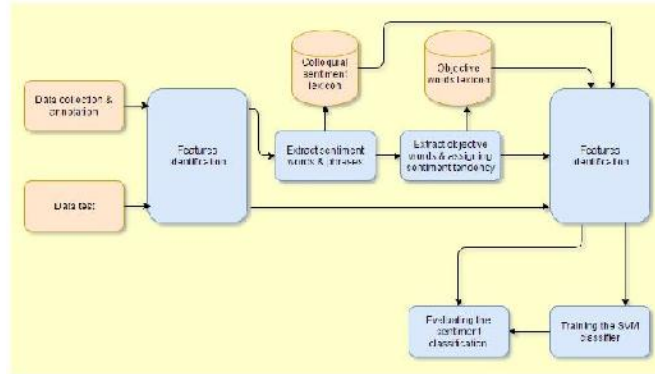

Figure 1: The framework of proposed approach.

## 4.2. DATA PREPROCESSING

In this phase, the dataset goes through a preprocessing of cleansing and preparation. The preprocessing includes correcting misspellings and removing repeated letters in words. We also removed punctuations, numerals, English words, and elongation. Emoticons are also removed because its usage was rare in the collected reviews. In this research, the process of stopwords removal is ignored since the stop-words may carry sentiment information and removing them affects the performance negatively. Next, a normalization process was applied to particular letters, for example the letters (أ، إ، آ) were converted to (ا), the letters (ى، ئ) were converted to (ي), the letter (ة) was converted to (ه), and finally the letter (ؤ) was converted to (و). The problem of the conjunctive particle (WA, و) was handled by applying a naive algorithm that simply removes the (و) from the beginning of any word containing more than three letters.

## 4.3. BUILDING COLLOQUIAL SENTIMENT LEXICON

In this phase, we manually built a sentimental lexicon consist of 3970 opinionated words and compound phrases. We extracted from our dataset 3367 opinionated words (adjectives, adverbs, nouns, and verbs) including colloquial and MSA. Words like English transliterations were included in the lexicon such as (نايس، لايك، جنتل). Furthermore, we extracted from our dataset 603 opinionated compound phrases that commonly used to express opinions. Compound phrases may indicate to specific groups of words, idioms, or speculations. Table 3 shows examples of dialectal compound phrases and individual words.

## 4.4. BUILDING OBJECTIVE WORDS LEXICON

The goal in this phase is to build a lexicon of objective words associated with its sentiment tendency to be used in the sentiment classification. In some studies such as [28, 37, 26, 27], the objective words are identified based on its polarity in the Senti Word Net, and then a new sentiment tendency is reassigned. In our work, since there is no such lexicon for colloquial Arabic language, the objective words will be extracted from our dataset. The basic observation shows that the review usually is expressed by both subjective and objective words. Knowing that, we already have built the lexicon of the subjective words in the previous module, so we can use it to identify the objective words. We create the lexicon of objective words through three steps. In the first step, we will use the conducted sentiment lexicon to eliminate the subjective words from our dataset, and the remaining words will be classified as objective words. Next, we will filter out the less informative words such stop-words, person names, and places from the objective words list. Finally, we identify the sentimental tendency of the objective words by adapting the method used in [26]. This method simply determines the sentiment tendency based on the ratio of its occurrences in either positive or negative reviews, as defined in Eq 1 and 2. For example, let us assume that the objective word (حكومة) which means government used more frequently in negative reviews. This word would be considered as it carries a negative tendency and labeled with a negative polarity, and vice versa.

Table 3: Examples of the colloquial words and compound phrases.

| Words & Phrases | Corresponding MSA | Polarity | Buckwalter | Gloss |
|---|---|---|---|---|
| مَناح | جيدون | Positive | mnAH | They are well |
| عجّة | إزدحام | Negative | Ejjp | Crowded |
| أنفلت | خُدعت | Negative | >nflmt | I have been deceived |
| سِرو فيه | يستحق | Positive | sErw fyh | Worthy |
| بطلوع الروح | بصعوبة | Negative | bTlwE AlrwH | Hardly |

$$pt(w_i) = \frac{no(pr_j, w_1)}{no(w_i)} \tag{1}$$

$$nt(w_i) = \frac{no(nr_j, w_1)}{no(w_i)} \tag{2}$$

Where pt indicates a positive tendency, nt is the negative tendency, pr is a positive review, nr is a negative review, w is an objective word and no (j,i) is the number of occurrences of i and j. However, some words may have an equal sentiment tendency value, or the difference of sentiment tendency is not enough to influence the process of polarity classification. For example, the total occurrences of an objective word I are 11 times of which 6 in positive reviews and 5 in negative reviews, the tendency values of positivity and negativity are 0.54 and 0.45 respectively. Although there is a difference between the tendency values of word i, it presents less information to learn from to classify the proper polarity. Therefore, we assigned a threshold value to categorize the objective words into three classes positive, negative or neutral. The sentimental tendency classification for the words is determined as Eq 3 shows.

$$t(w_i) = \begin{cases} Positive & pt(w_i) \geq 0.6 \\ Negative & nt(w_i) \geq 0.6 \\ Netural & otherwise \end{cases} \tag{3}$$

Where t indicates the tendency of objective word i. As a result, a lexicon of objective words was generated from our dataset. It consists of 7250 words out of 11388 words, of which 2941 were positive, 3297 were negative, and 1012 were neutral.

### 4.5. FEATURES IDENTIFICATION

In this section, we present the used features representation of the reviews in the dataset. In our approach,we used a set of sentiment features which include sentimental words, and the sentiment tendency of objective words. We also considered the sentimental compound phrases, the negation, and the length of reviews. More details about sentiment features are presented below:

• Sentiment words: This feature represents the words used to express positive or negative opinions. From this aspect, we have two features that represent the total number of positive and negative words in the review. To extract this feature, we built a colloquial sentiment lexicon which was introduced in Section 3.3.

• Sentiment tendency: This feature represents the total number of objective words that carry either a positive or negative tendency in the review. To extract this feature and the next feature, we built a colloquial objective words lexicon which was introduced in Section 3.4.

- Sentiment compound phrases: This type of phrases indicates grouping two or more words together to express the sentiment about an entity. For example, (همهم رضاك ,قول وفعل)which express a positive sentiment, and (فوق هذا كله ,كثير عليهم) which express a negative sentiment. Compound phrases also may refer to the idioms like (يا تاجر على عينك), or supplications like (اتقوا الله ,عنك له), (حسبي الله ونعم الوكيل). To extract this feature from the dataset, we used lexicons of positive and negative compound phrases containing 603 phrases. This feature is a binary feature value that set 1 for occurrence and -1 for absence.

- Negation words: The negation words often change the polarity of sentiment words to the opposite. For example, colloquial negation words such as (مو ، مش) can alter the sentiment orientation from positive to negative and vice versa. To handle this problem, we developed an algorithm that can change the polarity of sentiment words that follow any negation word. Wherever the negation word is found in the review, the algorithm will search any sentiment word within a scope of the three following words, and then the polarity will be reversed. After that, the appearances of negation words in the review will be counted and provided as a numeric feature.

• Review Length: The length of reviews in the dataset range from 2 to 159 words, the average is 23 words. Unlike the short review, longer reviews usually contain more details about the feeling of the opinion holder. However, not every word in the review is subjective. The length feature has shown a significant effect on sentiment classification in considerable amount of studies such as [38, 39, 40, 41].This feature is represented by the total number of the words in the review.

### 4.6. POLARITY CLASSIFICATION

Furthermore, we used the N-gram models to enable the classifier to learn at different levels of generality. N-gram is a classic model to represent features extracted from a text sequentially. The size of the sequence of words depends on the value of N, for example, a 2-gram is a two-word sequence of words like (بنصح تجرب) which mean (I recommend to try). 1-gram which is commonly known as unigram is the simplest model that provides a good coverage for the dataset, while 2-gram and 3-gram which are commonly known bigram and trigram respectively provide sufficient information to capture patterns of sentiment expressions. Therefore, as the size of the reviews in the dataset is long, we employed the N-gram with term occurrences weighting scheme to capture information might not be missed by the feature extractor such as negation words, contrary words, and sentiment compound phrases.

Next step after transforming the dataset into feature vector space is selecting the suitable learning classifier. In this work, we train a binary learning classifier to assign either positive or negative label to our dataset. We chose the SVM as our learning algorithm for classification, because the SVM classifier often yields a higher accuracy of performance than other machine learning algorithms as reported in most of the literatures [42, 13, 1, 43]. More specifically, we used LIBSVM [44] with linear kernel due to its efficiency.

## 5. EXPERIMENTAL RESULTS AND EVALUATION

This section presents the experimental results for the classifier used to classify the reviews into either positive or negative class. The data set used in the experiment was manually collected and annotated; we discussed the details in Section 3.1. To evaluate the performance of the proposed approach, we split our dataset into 85% as a training set and 15% as a testing set. The dataset includes reviews about diverse domains such as restaurants, shopping, fashion, education, entertainment, hotels, motors, and tourism. The average length of reviews is 23 words. To perform this experiment, we used Rapidminer. The Rapidminer is a software platform that includes a valuable set of machine learning algorithms and tools for data and text mining. We chose the following evaluation metrics: Accuracy, Precision, and Recall for evaluating the SVM classifier; see Eq 4, 5 and 6.

$$Accuracy = \frac{TP + TN}{TP + FP + TN + FN} \tag{4}$$

$$Precision = \frac{TP}{TP + FP} \tag{5}$$

$$Recall = \frac{TP}{TP + FN} \tag{6}$$

Where TP indicates a true positive which means the number of the inputs in data test that have been classified as positive when they are really belong to the positive class. TN indicates a true negative which means the number of the inputs in data test that have been classified as negative when they are really belong to the negative class. FP indicates a false positive which means the number of the inputs in data test that have been classified as positive when they are really belong to the negative class. FN indicates a false negative which means the number of the inputs in data test that have been classified as negative when they are really belong to the positive class.

The evaluation includes two phases. In the first phase, to determine the effectiveness of involving the objective words in the classification performance, we compared the performance between two baseline models and our model. The baseline models include a unigram model with term occurrences weighting scheme, and a model contains only basic sentiment features. We used the colloquial sentiment words lexicon to extract the basic sentiment features such as positive words (POw), negative words (NGw), positive compound phrases (POCP), negative compound phrases (NGCP), negation words (NEGw), and length of the review (LR). After that, we combined the sentiment objective words lexicon to extract the features such as positive tendency (POt) and negative tendency (NGt) to measure the effect these features on the classification; Table 4 shows the results.

The core of our work is that a sentimental review is usually made of more objective words than sentiment words, and these objective words may have sentiment tendency affecting the performance of polarity classification. The results in Table 4 show that the unigram baseline and the baseline model of basic sentiment features achieved accuracy with 89.9%, and 91.7% respectively. That means the lexicon-based features provided more sentiment information that can improve the performance of the classifier, especially the performance on positive instances. As a result, the precision improved by 3.5% compared to the unigram, because the false positives decreased. However, there was no improvement in the recall, as the false negatives are still high. The reason is that the negative reviews in our dataset contain less negative words and more objective words; thus the classifier tends to classify the negative reviews as positive especially when there are a few positive words appear.Back to the core of our work, it can be noted from the results that the addition of POt and NGt features has a significant improving effect on the accuracy performance of the classifier. In comparison with the unigram model, our approach improved the accuracy, precision and recall with 3.25%, 2.3%, and 4.4% respectively. As well as, compared to the other baseline, the improvement is in accuracy and recall with 1.45% and 4.4% respectively, while the precision decreased approximately 1%. As seen above, using the sentiment tendency of objective words was able to improve the overall performance and resolve the problem of less recall.

Table 4: Results of performance comparison between the proposed approach and the baseline models.

| Method | Features | Sentiment lexicon | Objective lexicon | Accuracy% | Precision% | Recall% |
|---|---|---|---|---|---|---|
| Baseline 1 | unigram | No | No | 89.9 | 91.7 | 87.68 |
| Baseline 2 | POw, NGw, NEGw, POCP, NGCP, LR | Yes | No | 91.7 | 95.2 | 87.68 |
| **Proposed** | **POw, NGw, NEGw, POCP, NGCP, LR, POt, NGt** | Yes | Yes | 93.15 | 94 | 92.12 |

The second phase includes four experiments that evaluate the performance of our proposed approach in conjunction with different n-grams and its combinations. Employing N-gram may improve the performance because it can store a higher degree of context than individual word. As our dataset includes long reviews, we assume that using n-grams will support in capturing more information about features like sentiment compound phrases, contrary word, and negation. We used three general representations of N-gram: unigram, bigram, and trigram. The SVM classifier was trained using sentiment feature set in conjunction with the occurrences of the N-gram models as the following: 1) sentiment feature set with unigram, 2) sentiment feature set with bigram, 3) sentiment feature set with a combination of unigram and bigram, 4) and finally sentiment feature set with a combination of unigram, bigram, and trigram. Table 5 shows the result of these experiments.

From the results in Table 5, we can clearly notice that each increment on the value of N in n-gram model combined with the proposed sentiment features has an improving effect on the performance of the SVM classifier. The results also show that combining the n-grams would perform even better, and it is noted that the more elements of n-grams in the combinations, the better performance. In other words, the combination of unigram and bigram yielded a higher performance than using n-grams separately with 95.35%. Also, adding the trigram to the combination outperformed all the above models by 95.6%. We suppose that this improvement because the classifier obtained more useful information about negation, and sentiment compound phrases by using n-gram features. To decide the optimal elements of the combination, we evaluated the addition of 4-gram to the combination, and we found that the addition of 4-gram has no improving impact on the performance, where its results are the same of the previous n-gram combination. Thus, the optimal model that would obtain the highest performance over our dataset is the combination of unigram, bigram, and trigram in conjunction with our proposed sentiment features.

Table 5: Results of combining proposed sentiment features with n-grams.

| Features | Accuracy% | Precision% | Recall% |
|---|---|---|---|
| Sentiment/unigram | 93.89 | 94.95 | 92.61 |
| Sentiment/bigram | 94.38 | 95 | 93.60 |
| Sentiment/trigram | 95.11 | 94.63 | 95.57 |
| Sentiment/unigram/bigram | 95.35 | 95.54 | 95.07 |
| **Sentiment/unigram/bigram/trigram** | **95.6** | **96.02** | **95.07** |
| Sentiment/unigram/bigram/trigram/4-gram | 95.6 | 96.02 | 95.07 |

The results obtained by evaluating the approach in the two phases with different feature sets indicate the use of sentiment tendency of objective words along with colloquial sentiment words can improve the sentiment classification performance. Additionally, using N-gram models and combining them together in conjunction with sentiment features also can improve the performance of colloquial sentiment classification. The highest results obtained by our classification approach show a significant improvement in accuracy, precision, and recall compared to the unigram baseline results which were reported in Table 1, by 5.7%, 4.3%, and 7.4% respectively, and compared to the baseline basic sentiment features the improvement is 3.9%, 0.82%, and 7.4% respectively.

## 6. CONCLUSION

In this paper, we have proposed a machine learning approach for colloquial Arabic sentiment analysis, specifically Jordanian colloquial reviews. The approach uses two new lexicons to extract the features: colloquial words sentiment lexicon, and objective words lexicon. Our dataset consists of 2730 reviews which have been manually collected and annotated. The annotated reviews dataset will be available publically for research purposes, as there is no one currently available. The core of our work is examining the effect of employing the objective words into the process of colloquial Arabic sentiment classification. After measuring the sentiment tendency of the objective words based on a particular threshold, the objective words lexicon was built. Then, we used it to extract two features: objective words with positive tendency, and objective words with negative tendency. These two features were incorporated with other sentiment features and n-grams to be fed to the SVM classifier. The experimental results show that the addition of objective words sentiment tendency has a significant improving effect on the sentiment classification compared to two baseline models.

## REFERENCES

[1]  H. Tang, S. Tan, and X. Cheng, "A survey on sentiment detection of reviews," Expert Systems with Applications, vol. 36, no. 7, pp. 10 760–10 773, 2009.

[2]  M. Hu and B. Liu, "Mining and summarizing customer reviews," in Proceedings of the tenth ACM SIGKDD international conference on Knowledge discovery and data mining. ACM, 2004, pp. 168–177.

[3]  B. Liu, "Sentiment analysis and subjectivity." Handbook of natural language processing, vol. 2, pp. 627–666, 2010.

[4]  W. Medhat, A. Hassan, and H. Korashy, "Sentiment analysis algorithms and applications: A survey," Ain Shams Engineering Journal, vol. 5, no. 4, pp. 1093–1113, 2014.

[5]  B. Liu, "Sentiment analysis and opinion mining," Synthesis lectures on human language technologies, vol. 5, no. 1, pp. 1–167, 2012.

[6]   M. Taboada, J. Brooke, M. To_loski, K. Voll, and M. Stede, "Lexicon-based methods for sentiment analysis," Computational linguistics, vol. 37, no. 2, pp. 267–307, 2011.

[7]   A. Kennedy and D. Inkpen, "Sentiment classi_cation of movie reviews using contextual valence shifters," Computational intelligence, vol. 22, no. 2, pp. 110–125, 2006.

[8]   M. Yang, W. Tu, Z. Lu, W. Yin, and K.-P. Chow, "Lcct: a semisupervised model for sentiment classi_- cation," in Human Language Technologies: The 2015 Annual Conference of the North American Chapter of the ACL. Association for Computational Linguistics (ACL)., 2015.

[9]   K. Shaalan, H. Bakr, and I. Ziedan, "Transferring egyptian colloquial dialect into modern standard arabic," in International Conference on Recent Advances in Natural Language Processing (RANLP–2007), Borovets, Bulgaria, 2007, pp. 525–529.

[10]  M. Abdul-Mageed, S. Kuebler, and M. Diab, "Samar: A system for subjectivity and sentiment analysis of social media arabic," in Proceedings of the 3rdWorkshop on Computational Approaches to Subjectivity and Sentiment Analysis (WASSA), ICC Jeju, Republic of Korea, 2012.

[11]  R. Duwairi, R. Marji, N. Sha'ban, and S. Rushaidat, "Sentiment analysis in arabic tweets," in Information and communication systems (icics), 2014 5th international conference on. IEEE, 2014, pp. 1–6.

[12]  H. ElSahar and S. R. El-Beltagy, "A fully automated approach for arabic slang lexicon extraction from microblogs," in International Conference on Intelligent Text Processing and Computational Linguistics. Springer, 2014, pp. 79–91.

[13]  B. Pang and L. Lee, "A sentimental education: Sentiment analysis using subjectivity summarization based on minimum cuts," in Proceedings of the 42nd annual meeting on Association for Computational Linguistics. Association for Computational Linguistics, 2004, p. 271.

[14]  N. A. Abdulla, N. A. Ahmed, M. A. Shehab, and M. Al-Ayyoub, "Arabic sentiment analysis: Lexiconbased and corpus-based," in Applied Electrical Engineering and Computing Technologies (AEECT), 2013 IEEE Jordan Conference on. IEEE, 2013, pp. 1–6.

[15]  R. M. Duwairi, "Sentiment analysis for dialectical arabic," in Information and Communication Systems (ICICS), 2015 6th International Conference on. IEEE, 2015, pp. 166–170.

[16]  J. Stokes and A. Gorman, "Encyclopedia of the peoples of africa and the middle east, 2010," Online Edition, The Safavid and Qajar dynasties, rulers in Iran from, vol. 1501, p. 707.

[17]  L. McLoughlin, Colloquial Arabic (Levantine). Routledge, 2009.

[18]  S. R. El-Beltagy and A. Ali, "Open issues in the sentiment analysis of arabic social media: A case study," in Innovations in information technology (iit), 2013 9th international conference on. IEEE, 2013, pp. 215–220.

[19]  P. McCarthy and C. Boonthum-Denecke, Applied natural language processing. Information Science Reference, 2011.

[20]  M. A. M. E. Ahmed, "Alarge-scale computational processor of the arabic morphology, and applications," Ph.D. dissertation, Faculty of Engineering, Cairo University Giza, Egypt, 2000.

[21]  N. Y. Habash, "Introduction to arabic natural language processing," Synthesis Lectures on Human Language Technologies, vol. 3, no. 1, pp. 1–187, 2010.

[22]  R. Hetzron, The Semitic Languages. Routledge, 2013.

[23]  A. Farghaly and K. Shaalan, "Arabic natural language processing: Challenges and solutions," ACM Transactions on Asian Language Information Processing (TALIP), vol. 8, no. 4, p. 14, 2009.

[24]  N. Habash and O. Rambow, "Magead: a morphological analyzer and generator for the arabic dialects," in Proceedings of the 21st International Conference on Computational Linguistics and the 44th annual meeting of the Association for Computational Linguistics. Association for Computational inguistics, 2006, pp. 681–688.

[25]  C. Miller, E. Al-Wer, D. Caubet, J. C. Watson et al., Arabic in the city: Issues in dialect contact and language variation. Routledge, 2007.

[26]  C. Hung and H.-K. Lin, "Using objective words in sentiwordnet to improve sentiment classi_cation for word of mouth," IEEE Intelligent Systems, vol. 28, no. 2, pp. 47–54, 2013.

[27]  Y. Kaewpitakkun, K. Shirai, and M. Mohd, "Sentiment lexicon interpolation and polarity estimation of objective and out-of-vocabulary words to improve sentiment classi_cation on microblogging." In PACLIC, 2014, pp. 204–213.

[28]  H. Amiri and T.-S. Chua, "Sentiment classi_cation using the meaning of words," in Workshops at the Twenty-Sixth AAAI Conference on Arti_cial Intelligence, 2012.

[29]  K. Ghag and K. Shah, "Sentit_df–sentiment classi_cation using relative term frequency inverse document frequency," International Journal of Advanced Computer Science & Applications, vol. 5, no. 2, 2014.

[30] H. S. Ibrahim, S. M. Abdou, and M. Gheith, "Sentiment analysis for modern standard arabic and colloquial," arXiv preprint arXiv:1505.03105, 2015.

[31] A. Shoukry and A. Rafea, "A hybrid approach for sentiment classi_cation of egyptian dialect tweets," in Arabic Computational Linguistics (ACLing), 2015 First International Conference on. IEEE, 2015, pp. 78–85.

[32] S. R. El-Beltagy, "Niletmrg at semeval-2016 task 7: Deriving prior polarities for arabic sentiment terms," Proceedings of SemEval, pp. 486–490, 2016.

[33] F. E.-z. El-taher, A. A. Hammouda, and S. Abdel-Mageid, "Automation of understanding textual contents in social networks," in Selected Topics in Mobile & Wireless Networking (MoWNeT), 2016 International Conference on. IEEE, 2016, pp. 1–7.

[34] A. M. Azmi and S. M. Alzanin, "Aara'–a system for mining the polarity of saudi public opinion through e-newspaper comments," Journal of Information Science, vol. 40, no. 3, pp. 398–410, 2014.

[35] J. B. Salamah and A. Elkhli_, "Microblogging opinion mining approach for kuwaiti dialect," Computing Technology and Information Management, vol. 1, no. 1, p. 9, 2016.

[36] A. Mourad and K. Darwish, "Subjectivity and sentiment analysis of modern standard arabic and Arabic microblogs," in Proceedings of the 4th workshop on computational approaches to subjectivity, sentiment and social media analysis, 2013, pp. 55–64.

[37] Z. Zhang, X. Li, and Y. Chen, "Deciphering word-of-mouth in social media: Text-based metrics of consumer reviews," ACM Transactions on Management Information Systems (TMIS), vol. 3, no. 1, p. 5, 2012.

[38] B. Pang, L. Lee, and S. Vaithyanathan, "Thumbs up?: sentiment classi_cation using machine learning techniques," in Proceedings of the ACL-02 conference on Empirical methods in natural language processing-Volume 10. Association for Computational Linguistics, 2002, pp. 79–86.

[39] S.-M. Kim, P. Pantel, T. Chklovski, and M. Pennacchiotti, "Automatically assessing review helpfulness," in Proceedings of the 2006 Conference on empirical methods in natural language processing. Association for Computational Linguistics, 2006, pp. 423–430.

[40] A. Ghose and P. G. Ipeirotis, "Designing novel review ranking systems: predicting the usefulness and impact of reviews," in Proceedings of the ninth international conference on Electronic commerce. ACM, 2007, pp. 303–310.

[41] J. Liu, Y. Cao, C.-Y. Lin, Y. Huang, and M. Zhou, "Low-quality product review detection in opinion summarization." in EMNLP-CoNLL, vol. 7, 2007, pp. 334–342.

[42] P. Beineke, T. Hastie, and S. Vaithyanathan, "The sentimental factor: Improving review classi_cation via human-provided information," in Proceedings of the 42nd annual meeting on association for computational linguistics. Association for Computational Linguistics, 2004, p. 263.

[43] A. Agarwal, B. Xie, I. Vovsha, O. Rambow, and R. Passonneau, "Sentiment analysis of twitter data," in Proceedings of the workshop on languages in social media. Association for Computational Linguistics, 2011, pp. 30–38.

[44] C.-C. Chang and C.-J. Lin, "Libsvm: a library for support vector machines," ACM Transactions on Intelligent Systems and Technology (TIST), vol. 2, no. 3, p. 27, 2011.